\documentclass[letterpaper, 10 pt, conference]{ieeeconf} 
\IEEEoverridecommandlockouts
\usepackage{cite}
\usepackage{amsmath,amssymb,amsfonts}
\usepackage{algorithmic}
\usepackage{graphicx}
\usepackage{float}
\usepackage{caption}
\usepackage{textcomp}
\usepackage{xcolor}
\usepackage{cuted}
\usepackage{booktabs}
\usepackage{array}
\usepackage{multirow}

\def\BibTeX{{\rm B\kern-.05em{\sc i\kern-.025em b}\kern-.08em
    T\kern-.1667em\lower.7ex\hbox{E}\kern-.125emX}}

\begin{document}

\title{\fontsize{23pt}{26pt}\selectfont Control Without Control: Defining Implicit Interaction Paradigms for Autonomous Assistive Robots\\}

% \author{\IEEEauthorblockN{1\textsuperscript{st} Janavi Gupta}
% \IEEEauthorblockA{\textit{dept. name of organization (of Aff.)} \\
% \textit{name of organization (of Aff.)}\\
% City, Country \\
% email address or ORCID}
% \and
% \IEEEauthorblockN{2\textsuperscript{nd} Given Name Surname}
% \IEEEauthorblockA{\textit{dept. name of organization (of Aff.)} \\
% \textit{name of organization (of Aff.)}\\
% City, Country \\
% email address or ORCID}
% \and
% \IEEEauthorblockN{3\textsuperscript{rd} Given Name Surname}
% \IEEEauthorblockA{\textit{dept. name of organization (of Aff.)} \\
% \textit{name of organization (of Aff.)}\\
% City, Country \\
% email address or ORCID}
% \and
% \IEEEauthorblockN{4\textsuperscript{th} Given Name Surname}
% \IEEEauthorblockA{\textit{dept. name of organization (of Aff.)} \\
% \textit{name of organization (of Aff.)}\\
% City, Country \\
% email address or ORCID}
% \and
% \IEEEauthorblockN{5\textsuperscript{th} Given Name Surname}
% \IEEEauthorblockA{\textit{dept. name of organization (of Aff.)} \\
% \textit{name of organization (of Aff.)}\\
% City, Country \\
% email address or ORCID}
% \and
% \IEEEauthorblockN{6\textsuperscript{th} Given Name Surname}
% \IEEEauthorblockA{\textit{dept. name of organization (of Aff.)} \\
% \textit{name of organization (of Aff.)}\\
% City, Country \\
% email address or ORCID}
% }
\author{Janavi Gupta$^{1*}$, Kavya Puthuveetil$^{1*}$, Dimitra Tsakona$^{2}$, Akhil Padmanabha$^{1}$,\\Yiannis Demiris$^{2}$, and Zackory Erickson$^{1}$ % <-this % stops a space
\thanks{}
\thanks{$^{*}$These authors contributed equally.}
\thanks{$^{1}$Robotics Institute, Carnegie Mellon University, Pittsburgh, PA, USA {\tt janavig@andrew.cmu.edu, kavya@cmu.edu}}
\thanks{$^{2}$Imperial College London, London, UK}%
\thanks{This work was supported by the National Science Foundation Graduate Research Fellowship Program under Grant No. DGE2140739}
% \thanks{All authors are with the Robotics Institute, Carnegie Mellon University, Pittsburgh, PA, USA {\tt janavig@andrew.cmu.edu, kavya@cmu.edu}}
}%

\maketitle
 
\begin{abstract}
Assistive robotic systems have shown growing potential to improve the quality of life of those with disabilities. As researchers explore the automation of various caregiving tasks, considerations for how the technology can still preserve the user's sense of control become paramount to ensuring that robotic systems are aligned with fundamental user needs and motivations. In this work, we present two previously developed systems as design cases through which to explore an interaction paradigm that we call \textit{implicit control}, where the behavior of an autonomous robot is modified based on users' natural behavioral cues, instead of some direct input. Our selected design cases, unlike systems in past work, specifically probe users' perception of the interaction. We find, from a new thematic analysis of qualitative feedback on both cases, that designing for effective implicit control enables both a reduction in perceived workload and the preservation of the users' sense of control through the system's intuitiveness and responsiveness, contextual awareness, and ability to adapt to preferences. We further derive a set of core guidelines for designers in deciding when and how to apply implicit interaction paradigms for their assistive applications.

\end{abstract}

\section{Introduction}
Motor impairments can drastically limit a person’s ability to perform Activities of Daily Living (ADLs), which include fundamental self-care tasks such as bathing, eating, and dressing, as well as Instrumental Activities of Daily Living (iADLs) such as home management, cooking, and cleaning. In the United States alone, approximately 5 million individuals live with partial or total paralysis resulting from stroke, spinal cord injury, neurodegenerative disorders, or progressive muscular diseases~\cite{armour2016prevalence}. Many individuals with reduced motor function must rely heavily on caregivers for routine tasks such as scratching an itch, eating, opening a door, or getting dressed~\cite{dijkers2005quality, fernhall2008health, mercier2001impact, sakakibara2009systematic, wang2023one}. This can significantly affect independence, sense of agency, and overall quality of life.

\begin{figure}[h]
    \centering
    % \vspace{-2.3cm}
    \includegraphics[width=\linewidth, trim={8.5cm 0cm 8.5cm 0cm}, clip]{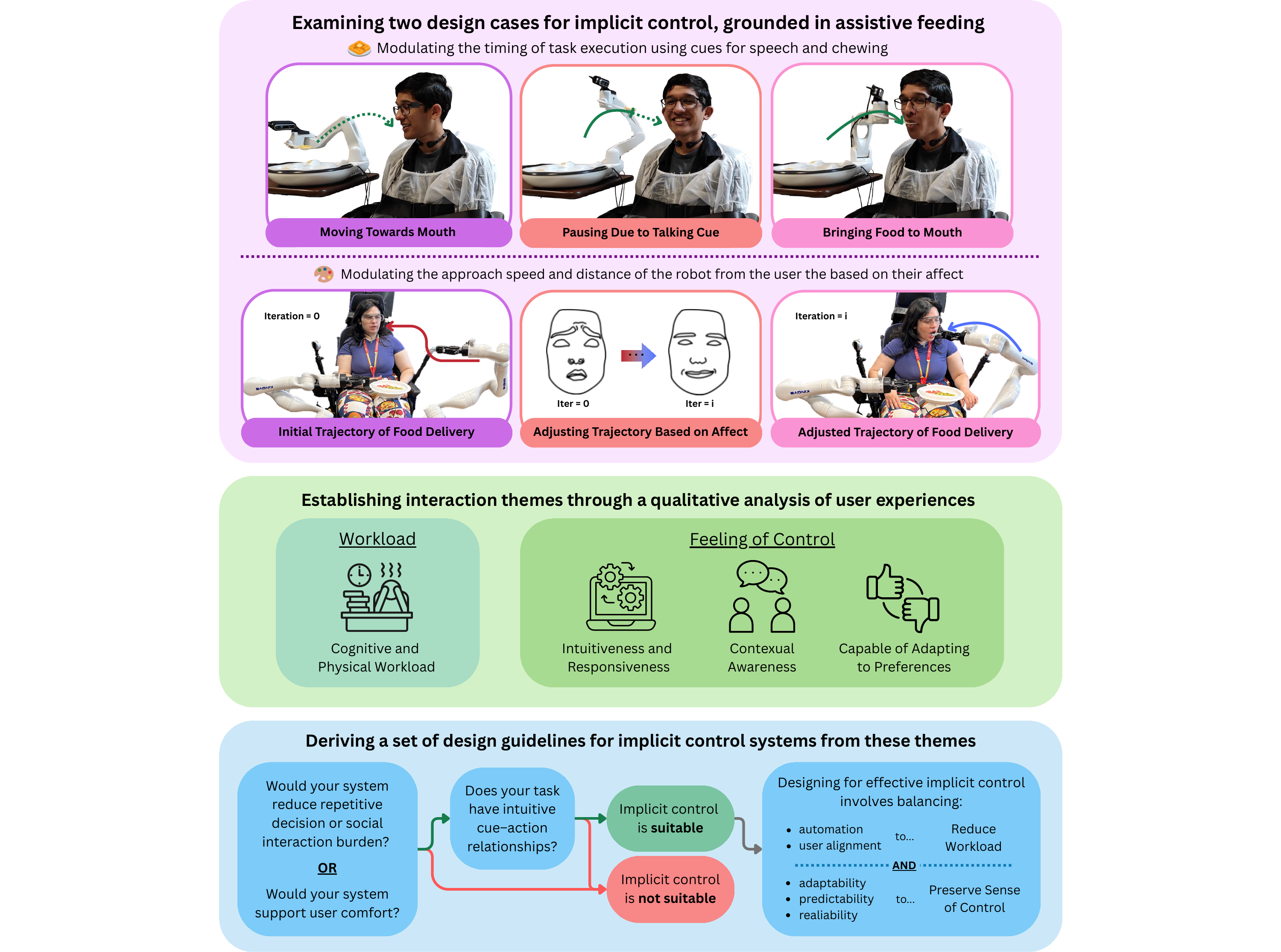}
    \captionof{figure}{\textit{Top:} We present two assistive systems as design cases for implicit control. \textit{Middle:} We perform a thematic analysis of users' qualitative feedback on interactions with the systems and derive a set of core themes. \textit{Bottom:} From these themes, we establish a set of design guidelines.}
    \label{fig:teaser}
    \vspace{-0.7cm}
\end{figure}

Assistive robots have shown growing promise in helping individuals with motor impairments perform these daily tasks~\cite{bunketorp2021surgical, collinger2013high, maceira2019wearable, osuagwu2018clinical, son2022biomechanical}. Advances in perception, manipulation, and motion planning have expanded the technical capabilities of such systems, enabling robots to perform increasingly complex assistive behaviors. However, the success of assistive robotics depends not only on what the robot can do, but also on how people interact with it. In many assistive scenarios, users must coordinate closely with the robot during everyday activities, making the design of interaction and control mechanisms central to whether these systems feel usable, comfortable, and empowering.

Thus, a central challenge is determining how users guide robot behavior. Many assistive systems rely on active control, in which users explicitly trigger robot actions through a control interface~\cite{fehr2000adequacy, kibria2005speech, stiefelhagen2004natural}. These approaches provide clarity and predictability, allowing users to intentionally direct the robot’s behavior. Other systems incorporate shared autonomy, where the robot assists with execution while the user provides higher-level guidance~\cite{padmanabha2024independence, admoni2016predicting}. Still other systems take things a step further, with the robot operating completely autonomously. While these systems may effectively alleviate nearly all cognitive and physical burden from the user, an important interaction question, grounded in the fundamental goal in assistive technology design to preserve users' sense of agency~\cite{zhang2022learning, sun2024force}, emerges: \textbf{how can users maintain a sense of control when the robot is independently making decisions about when and how to act?}

To this end, we propose an interaction-centric reframing of autonomous systems as a paradigm we term \textit{implicit control}, in which the robot acts autonomously but regulates its behavior based on some implicit cues from the user. The core principle is that designing autonomous assistive systems to respond to natural signals from the user may enable users to feel in control, even if they don't actually have any explicit control input. Many existing autonomous assistive systems can, in principle, be viewed as instances of implicit control in that the robot acts based on indirect cues from the user~\cite{omer2025real, zhang2025mindeye, wang2022implicit, wang2023one}. However, most of this prior work has largely focused on enabling autonomy and evaluating task performance, rather than analyzing how users perceive interactions with these systems and what design features support such interactions, particularly with respect to sense of control. From our design analysis summarized in Fig.~\ref{fig:teaser}, we present a set of guidelines to address this gap and support researchers, engineers, and designers in developing autonomous assistive systems that both reduce the interaction burden and preserve users' sense of control.

In this paper, we perform a design analysis in which: 
\begin{itemize}
    \item We examine implicit control through two design cases for systems that modulate robot behavior based on indirect user behavioral cues.
    \item We conduct a qualitative analysis of user experiences with these systems and identify a recurring set of interaction themes that show how implicit control reduces workload while preserving the user's sense of control.
    \item Based on these analyses, we derive a set of design guidelines for embedding implicit control into assistive robotic systems.
\end{itemize}
 
\section{Related Work}
\subsection{Interaction Paradigms}
Traditional frameworks for understanding control in robotics have largely emerged from teleoperation, supervisory control, and autonomy, where the focus is on system architecture and task performance. Consequently, assistive robotics has also drawn from these frameworks, typically describing control in terms of how much decision-making power the robot has relative to the human (e.g., levels of automation), with arbitration algorithms determining how human input and autonomous assistance are combined~\cite{parasuraman2000model,sheridan2002humans,kaber2004loa}. In this view, interaction paradigms are typically described along a spectrum from direct teleoperation to shared control and full autonomy, with the key design variables commonly being the robot’s capabilities, the human’s input bandwidth, and arbitration policies that blend the human and robot's control~\cite{gopinath2016human, javdani2018shared, dragan2013policy}. 

However, this framing fails to consider a user-facing question that is central in assistive contexts: how can users feel in control, safe, and comfortable, even if they are not issuing explicit commands? Previous work has established that users with mobility impairments do not uniformly prefer ``more autonomy,'' and preferences depend on comfort, predictability, and how the robot’s behavior fits the interaction context \cite{bhattacharjee2020}, further underscoring the need to consider such factors, in addition to how users can feel in control, when designing autonomous systems. There is growing interest across human-robot interaction, and especially in the assistive space, in designing autonomous systems that support users' sense of agency~\cite{bhattacharjee2020,selvaggio2021autonomyphri,argall2018rehabautonomy}.

\subsection{Implicit Control}
Many previous works in the assistive space have developed autonomous technologies that modulate their behavior based on indirect signals from the user, with the most common being methods that perform a task based on the human's pose, generally estimated with vision~\cite{zhang2022learning, wang2023one}. Involuntary physiological signals or cues, like error-related potentials from EEG~\cite{wang2022implicit, omer2025real}, eye gaze~\cite{zhang2025mindeye, belcamino2024gaze}, and stress-related measures (e.g., heart rate, galvanic skin response, skin temperature, wincing)~\cite{messeri2021human, mower2007investigating, grice2012wouse}, have also been widely explored for informing robot actions. In the latter case, more consideration has been given to features that impact the user's perception of the robot, instead of just evaluating pure task performance. However, gaps remain across the spectrum of work in this area, which motivates our interaction-centric reframing of these types of autonomous systems as \textit{implicit control}. The implicit control systems we examine in this work~\cite{padmanabha2025waffle, tsakona2025continuous}, in contrast to much of the existing literature, do include user reflections on the interaction as key aspects of the evaluation, allowing us to investigate the design features that make them effective and capable of maintaining users' sense of control, which we intrepet as the user’s perception that their behavior meaningfully influences the robot.

% In this work, we examine two implicit control systems~\cite{padmanabha2025waffle, tsakona2025continuous} that do include user reflections on the interaction as key aspects of the evaluation in order to elucidate what features of these systems allow them to be effective and maintain users' sense of control.

\section{Methods}
In this section, we first summarize two implicit control systems, WAFFLE~\cite{padmanabha2025waffle}, which coordinates task timing, and CRAFTT~\cite{tsakona2025continuous}, which responds to user affect. The two systems were developed independently of each other, with WAFFLE coming from a team of researchers at Carnegie Mellon University with collaborators at Cornell University, and CRAFTT originating from a team at Imperial College London; a subset of authors on this paper contributed to these prior works. We highlight the aspects of these systems' design and evaluation most relevant to understanding their relevance as prototypical cases of implicit control. We then describe our procedure for a new thematic analysis of qualitative user reflections on both these systems. While the original papers evaluated system performance and task-specific usability, this paper uniquely extracts generalized interaction themes that span across disparate sensing modalities.

\subsection{System Overview}
\subsubsection{WAFFLE}
WAFFLE is a wearable-based robot-assisted feeding system that estimates when a user is ready for the next bite using implicit behavioral cues. The user wears glasses equipped with an inertial measurement unit (IMU) and a contact microphone, which capture head motion and vibrations associated with chewing and speech directly from the user in a way that is robust to occlusion and environmental noise. These signals are processed in real time to distinguish between when the user is occupied and when they are ready for the next bite. We deploy WAFFLE on the Obi feeding robot~\cite{Obi}, which follows a predefined trajectory from a staging position toward the user’s mouth. WAFFLE modulates the robot’s motion along this trajectory by pausing or advancing based on the inferred user state, continuously coupling the user’s natural activity to the robot's behavior.

We evaluated WAFFLE in a user study with 15 participants without motor impairments and 2 participants with motor impairments. Participants experienced three bite-timing methods: (a) wearable-based implicit timing (WAFFLE), (b) a mouth-open method requiring explicit user signaling (active control), and (c) a fixed-interval method with no user input (fully autonomous timing). The study was conducted in both individual and social dining settings. After each condition, participants completed Likert-scale questionnaires and provided open-ended feedback. The complete quantitative results are reported in our prior work~\cite{padmanabha2025waffle}.

\subsubsection{CRAFTT}
CRAFTT is a framework, which we demonstrate in a robot-assisted feeding scenario, that aims to maximize user comfort while maintaining task efficiency and reducing physical burden by continuously adapting the robot's behavior based on implicit affective cues from the user. The system uses RGBD video to capture facial micro-expressions, head motion, and gaze-related features, which serve as proxies for user comfort and stress. Based on deviations in affect from a user-specific baseline, CRAFTT iteratively updates the robot's approach velocity and final delivery distance by solving, in real-time, a multi-objective optimization problem that balances user comfort, task efficiency, and physical strain. Thus, CRAFTT enables the robot to respond continuously to the user without requiring explicit input. We deployed CRAFTT on the Blueberry dual-arm wheelchair robotic platform, where one arm stabilizes a plate of food and the other feeds the user.

In a study with 26 participants without motor impairments, we compared user interactions with CRAFTT, our adaptive system, to a non-adaptive baseline. Our evaluation combined objective measures of user comfort and task efficiency with subjective assessments of trust, reliance intention, perceived safety, fluency, technology acceptance, and cognitive load. 
% Results show that adaptation improves trust, reliance, acceptance, and interaction fluency while maintaining efficiency. 
The full experimental details and quantitative results are reported in our prior work~\cite{tsakona2025continuous}.

\subsection{Design Motivation}

In WAFFLE, we were primarily concerned with designing a system that could coordinate bite timing without explicit prompting since such methods (e.g., button press, facial gesture) place additional interactional burden on the user and can be disruptive in social contexts, requiring users to divide their attention between participating in the conversation and managing the robot. CRAFTT, on the other hand, considered how systems, especially those assisting with high-intimacy or risky tasks like feeding, must account for the user’s affective state and adjust their behavior accordingly to support a balanced sense of agency and comfort throughout the interaction. Despite the completely independent development of these two systems for two separate contexts, we see that both WAFFLE and CRAFTT converge on a common set of key design choices. Specifically, these systems:

\begin{itemize}
    \item rely on implicit cues to adjust robot behavior, instead of relying on explicit input, to \textit{reduce cognitive and physical effort}.
    \item modulate robot behavior based on real-time feedback, instead of simply triggering an open-loop routine, to enable continuous \textit{responsiveness} to the user.
\end{itemize}

We contend that these qualities, namely that the robot's autonomous behavior is grounded in cues made by the user, make these prototypical implicit control systems. We explore how this user-centered approach to robot autonomy shapes perceptions of interaction with the robot in a thematic analysis of participant feedback to both WAFFLE and CRAFTT.

\subsection{Thematic Analysis Procedure}
As WAFFLE and CRAFTT were developed independently, their respective studies used different qualitative items and warranted different initial procedures for their analysis, before finally combining to arrive at a unified set of themes.

\subsubsection{WAFFLE}
Following the user study in WAFFLE, participants were asked to respond to a battery of qualitative items, of which we down-selected to the four most relevant to our exploration of implicit control for this work. Specifically, we focus on questions that ask participants to reflect on their sense of control when interacting with the robot, the appropriateness of the robot's bite timing, similarities or differences between the robot and a human caregiver, and contextual differences between social and individual dining:
\begin{enumerate}
    \item Why did you prefer your highest-ranked method? Please elaborate and compare the methods for both the individual and social settings.
    \item Which methods did you feel in control while using? Why or why not?
    \item Which method had the most appropriate bite timing? Elaborate on why it had the most appropriate timing and compare it to the other methods.
    \item Which method was closest to how a caregiver, or you, would perform this task? Please elaborate.
\end{enumerate}
 
In performing our thematic analysis of participant responses, we took a hybrid inductive–deductive approach with coding done by a single annotator from our research team. We started with a deductive round of analysis, segmenting participant responses based on their alignment with hypothesized thematic constructs, corresponding to the following four codes: \textit{feeling of control, natural cues, social compatibility, cognitive and physical workload}. Through this process, we observed how participants’ responses on their interaction with WAFFLE converged to reflections on either the effort required or the degree of control they felt. We followed with three iterative rounds of inductive analysis in which codes were merged and refined when themes had overlapping semantic meaning. As a result, our final core themes emerged: \textbf{workload} and \textbf{sense of control}, with individual sub-themes around \textit{seamlessness}, \textit{predictability}, and \textit{contextual awareness} surfacing as key features that influence the latter. 

\subsubsection{CRAFTT}
In our experiments with CRAFTT, we asked each participant the following at the end of their session: ``Assuming you needed the robot's assistance, which robot behavior [adaptive (CRAFTT) vs. non-adaptive (baseline)] would you choose to continue interacting with and why?'' We center our examination of implicit control in this work on responses to this question.

Similar to WAFFLE, the thematic analysis of responses to CRAFTT was also done by a single annotator using a hybrid inductive-deductive approach, with the deductive round starting with three codes, \textit{perceived responsiveness}, \textit{intuitive interaction}, and \textit{personalization}, based on the work's hypothesized thematic constructs. Following this, the additional theme of \textit{effort redistribution} emerged, and a final round of deductive coding was performed to apply this fourth code.

\subsubsection{Unifying Themes}
Following the two independent thematic analyses of WAFFLE and CRAFTT, we observed that responses aligned with the themes of ``workload'' and ``effort redistribution'' from WAFFLE and CRAFTT, respectively, shared semantic similarities, as did those from the themes ``seamlessness'' and ``predictability'' (WAFFLE) and ``intuitive interaction'' and ``perceived responsiveness'' (CRAFTT). In contrast, the responses aligned with the theme of ``contextual awareness'' from WAFFLE and ``personalization'' from CRAFTT did not share any commonality with those from other themes or with each other.

Based on how the themes converged semantically between systems, we synthesized the following unified thematic structure that bridges the two: \textbf{workload} and \textbf{sense of control}, with individual sub-themes for the latter around \textit{intuitiveness and responsiveness}, \textit{personalization}, and \textit{contextual awareness}. We then conducted a final round of coding, applying this shared set of themes across participant responses from both WAFFLE and CRAFTT to enable direct comparison and analysis. The results of this final round are discussed in Sec.~\ref{sec:results}.

\begin{figure*}[t!]
  % \vspace{-0.1cm}
  \centering
  \includegraphics[width = 0.85\textwidth]{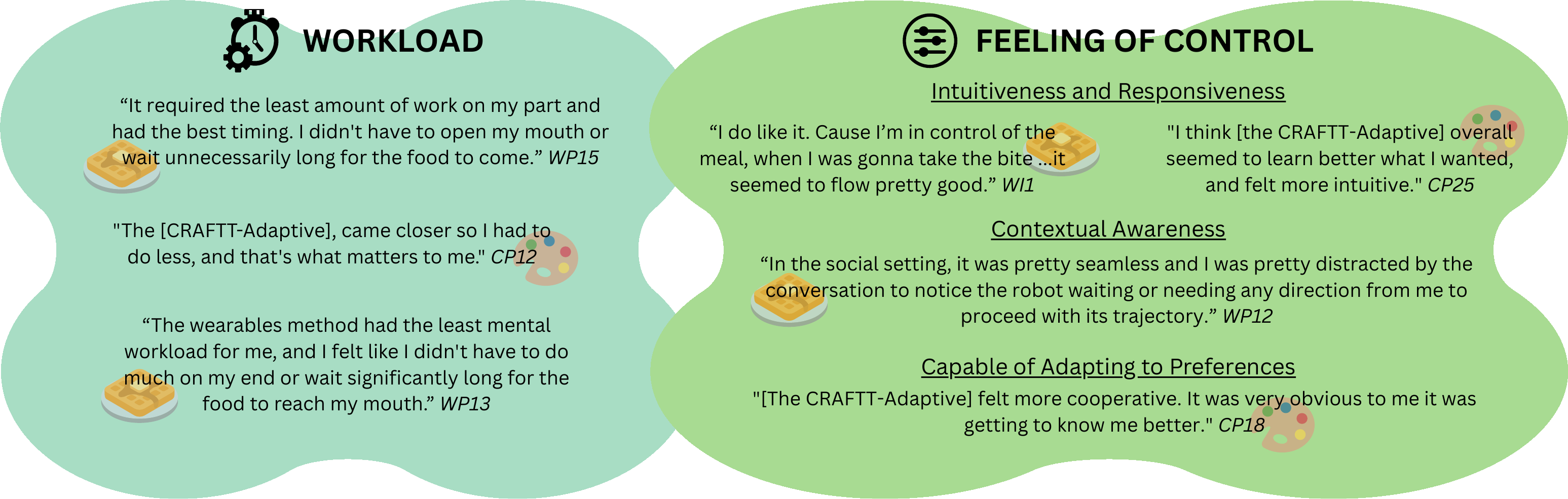}
  % \vspace{-0.3cm}
  \caption{A subset of quotes from both the WAFFLE and CRAFTT studies, presented and organized according to the categories outlined in our thematic analysis.}
  \vspace{-0.6cm}
  \label{fig:quotes}
\end{figure*}

\section{Results}
In this section, we use WP and WI to denote quotes from participants in the WAFFLE study without and with impairments, respectively, while we use CP for participants, none of whom had impairments, in the CRAFT study. Throughout the analysis, we report the number of participant comments aligned with a given theme as a proportion of all \textit{relevant} comments, instead of the total number of participants, since not all participants touched on every aspect of the interaction in their qualitative responses. A summary of the interaction themes extracted from our thematic analysis, along with representative quotes for each, is presented in Fig.~\ref{fig:quotes}. 

\label{sec:results}
\subsection{Workload} 
Across both systems, interactions were described as reducing cognitive and physical workload by removing the need for explicit coordination, with all 7/7 relevant participant responses from the WAFFLE study and 4/4 from the CRAFTT study in agreement. For WAFFLE, participants emphasized that they did not need to think about signaling for each bite or interrupt their activity to issue commands. As WP2 explained, the wearable system required ``the least amount of thinking.'' This sentiment was corroborated by WP9, who noted that WAFFLE ``did not require me to actively do anything.'' Participants attributed this reduction in effort to the elimination of repeated micro-decisions, allowing them to focus on eating or conversation while the system inferred appropriate timing.

Similarly, in the CRAFTT study, participants described a shift in effort from themselves to the robot. When interacting with the non-adaptive baseline, users reported that they were compensating for the robot’s fixed actions. For example, CP11 noted that they were ``putting in all the effort,'' while CP8 described ``trying too hard'' to maintain coordination. The adaptive system, on the other hand, redistributed this burden, enabling a more balanced interaction where the system adjusted to the user rather than requiring the user to adapt.

\textbf{Taken together, these findings suggest that simply automating a task does not guarantee perfect redistribution of workload away from the user to the robot; if the system is incapable of adapting to the user, the user will compensate through increased effort or attention}. In contrast, implicit control systems, which do adapt to the user, shift the coordination burden to the robot, thereby reducing user workload.
\textbf{That being said, we observed that a successful reduction in workload was conditional on the adaptation being aligned with user expectations.} In the WAFFLE study, misalignment was associated with increased monitoring effort. As WP12 noted, the system ``sometimes didn't pick up on the fact that I was still chewing,'' requiring vigilance to ensure appropriate behavior.

\subsection{Feeling of Control}
Participants’ self-described sense of control over the robot was more complex than simply being able to issue a direct command. Instead, \textbf{the feeling of control emerged as a multi-dimensional construct shaped by intuitiveness and responsiveness, contextual awareness, and the capability to adapt to preferences throughout the interaction}.

\subsubsection{Intuitiveness and Responsiveness}
11/15 relevant participant responses in the WAFFLE study and 13/15 in the CRAFTT study described feeling in control when the robot’s behavior appeared to respond to their actions and internal state, often in ways that felt intuitive and required little explicit reasoning.

For WAFFLE, participants reported that they felt the robot advanced only when they were ready, even without having to explicitly prompt it to wait. As WI1 explained, ``I'm in control of the meal... it stopped when I wanted to,'' while WP14 noted that the robot ``waited for some sort of cue from me.'' Similarly, participants described the interaction as seamless, with the robot yielding when they were chewing or speaking and advancing when they were ready. These experiences suggest that alignment between the user’s natural behavior and the robot’s actions can evoke a strong sense of control even without explicit commands.

A similar pattern emerged in responses to the CRAFTT system, where participants frequently described the robot as ``understanding'' them or ``responding'' to their behavior. For example, CP4 noted that the robot ``seems to understand me and when I react, it just changes how it acts,'' while CP20 stated that it ``respected my boundaries.'' This is particularly notable given that participants were blinded to the experimental conditions they were interacting with during the study. Despite not being informed that the CRAFTT system was performing any adaptation, users detected that it was, suggesting that the degree to which the robot's behavior changed in response to the user was perceptually salient and sufficient to convey the user's influence over the system. 

Responsiveness also contributed to the intuitiveness of the interaction. Many participants preferred the adaptive behavior without being able to clearly articulate why, describing it as a ``gut feeling'' (CP13) or as ``intuitive'' (CP25). In some cases, participants reported feeling able to influence the robot, as CP27 noted, ``I really felt I could stop it with my reactions,'' but lacked a complete picture of how. \textbf{These findings suggest that users don't necessarily need to be able to accurately reason about how exactly the system works for them to feel in control; a rough intuition of how the robot responds to their implicit signals may be sufficient.} Regardless, designers must still be thoughtful about execution. In WAFFLE, we observed how errors, such as premature bite delivery, or certain constraints on the user, like how some felt they needed to remain ``unnaturally still'', can result in misalignments with users' expectations that hinder perception of the robot as intuitive, and, therefore, diminish their sense of control.

Although most participants across the studies preferred implicit control, not all did. In the WAFFLE study, a few participants preferred the mouth-open baseline because it provided a clear and direct mechanism for controlling the robot. As WP3 explained, it ``relied on an explicit cue,'' while WP5 noted that delaying the robot required ``coming up with something'' to influence its behavior. For these participants, a sense of control required explicit, intentional signaling. Similarly, in the CRAFTT study, some participants preferred non-adaptive behavior due to its consistency. These responses highlight a tension: \textbf{while responsiveness can enhance intuitive interaction, it can reduce predictability, suggesting that effective implicit control must balance adaptability with stable and interpretable behavior.}

\subsubsection{Contextual Awareness}
%% WAFFLE
We saw that the theme of contextual awareness emerged uniquely from the conditions of the WAFFLE study, with no associated responses from CRAFTT, primarily given the social dining setting included in the system evaluation. 4/5 relevant responses from WAFFLE participants without impairments and 1/2 from those with impairments described the wearable method as the least disruptive to conversation. WI1 emphasized that it allowed them to ``have a conversation and eat at the same time.'' WP9 noted it ``did not require me [to interrupt] the conversation,'' and WP15 appreciated that the robot stayed still during speech so that ``it wouldn’t make the conversation awkward.'' \textbf{Here, control extended beyond robot timing to include control over one’s environment.} Participants felt most in control when the system supported conversational flow and aligned with social norms.

\subsubsection{Capable of Adapting to Preferences}
%% CRAFTT
As WAFFLE did not involve any specific user preferences, acting more based on a binary decision of whether or not the user was ready for the next bite, this theme only applied to responses to CRAFTT. Participants in the CRAFTT study highlighted preferences for specific behavioral characteristics of the robot, which were often contradictory across individuals, with positive feedback from all 9/9 participants with relevant responses. For example, CP2 preferred the adaptive robot because ``it was much faster,'' emphasizing efficiency, while CP1 directly expressed preference for the adaptive robot because ``it moved slower, like I wanted''. Similarly, CP12 favored the adaptive system because ``it came closer so I had to do less'', whereas CP11 reported that the robot ultimately behaved as desired by ``move[ing] slower and stop[ping] further [away]''. These contrasting preferences suggest that users do not share a single notion of optimal behavior, but instead value different execution characteristics depending on their comfort, priorities, and sensitivities. Notably, these differences were not merely perceptual: the system adapted to each user and converged to distinct execution strategies across participants. \textbf{Participants frequently interpreted these personalized adjustments as the robot ``understanding'' them}, as mentioned by CP1 and CP4, \textbf{reinforcing their sense of control over the interaction.}

\section{Design Guidelines for Implicit Control}
Based on the design tensions identified above, as well as foundational principles of assistive technology~\cite{ortiz2023assessing, kinney2016measuring}, we derive a set of design guidelines, framed as a decision tree, intended to support the development of implicit control systems in assistive robotics. First, we identify when implicit control is an appropriate interaction paradigm, namely when the task 1) is \textit{burdensome to the user} or 2) \textit{necessitates accounting for comfort}, and when 3) \textit{there are natural, interpretable user signals that can guide robot behavior}. We then highlight conditions for designing an effective implicit control system, namely that it balances 1) \textit{task automation} and \textit{alignment to user expectations} to \textbf{reduce workload}, and 2) \textit{adaptability}, \textit{predictability}, and \textit{reliability} to \textbf{preserve the user's sense of control}. 
% Together, these guideline articulate interaction-level principles that designers can apply when developing assistive systems that respond to implicit user behavior.

\subsection{When to Use Implicit Control}
% Our findings suggest that implicit control is best applied when 1) the task is significantly burdensome to the user, or 2) the context of the task necessitates accounting for comfort, \textit{and} 3) there is some natural signal from the user that is both useful and interpretable. We describe these conditions in greater depth as follows.
% We identify three conditions under which passive control is particularly well suited: when cue–action relationships are predictable, when interaction involves repetitive decisions, and when explicit signaling would create social or contextual strain.
\subsubsection{To Alleviate the User of Burden}

As is the case when considering autonomy in any context, designers must first contend with why the task should be automated in the first place. In the assistive realm, we presume that the most pressing or important tasks to consider automating are those that cause some significant burden or strain on the user. One of the forms of burden that WAFFLE particularly attempted to mitigate was that from repeated decision-making and explicit cuing. Participants valued not having to continuously signal readiness for each bite, unlike with methods that required them to perform the same gesture or make the same timing decision over and over. 
% Even if these decisions occur relatively infrequently, they may still create an interaction burden when they interrupt a user’s natural workflow.
In this case, the implicit control system allowed users to focus on the activity itself rather than on managing the robot. 
WAFFLE also considered another type of burden: social interaction burden. Participants noted that WAFFLE allowed them to engage in conversation without explicitly managing the robot, preserving the natural flow of social interaction during meals. Participants appreciated that the robot paused during conversation and resumed when it lapsed, unlike baseline methods that required users to monitor for the right time to explicitly cue for a bite, or which had a fixed timer that could interrupt at any point.

WAFFLE's ability to time bites based on the user's natural behavior eliminated the interaction burden that would otherwise interrupt eating or conversation. By responding to ongoing behavior, the system allowed them to focus on the activity itself rather than on managing the robot. Therefore, \textbf{designers should consider implicit control for tasks that involve recurring coordination decisions}, such as pacing, pausing, or advancing within an ongoing activity, \textbf{or in socially embedded activities, among any other forms of interaction burden a user may encounter.}

\subsubsection{To Support User Comfort}
Implicit control can also be considered in scenarios where interactions with the robot involve an inherent sense of vulnerability and perceived risk on the user's part, which is often the case in physical assistance. Our findings with CRAFTT indicate that users' sense of control can be supported by the robot not only modulating \textit{what} it does or \textit{when}, but also \textit{how} it does it, particularly with respect to adjustments that make users feel more comfortable. Implicit control enables the system to infer and adapt to fine-grained variations in user affect, supporting more personalized and responsive interaction than may be possible through direct input, since preferences and comfort can be difficult to articulate, even more so in real-time. Thus, \textbf{designers may consider applying implicit control for tasks where execution dynamics critically influence user comfort or safety, particularly in close-proximity and high-intimacy interactions.}

\subsubsection{There are Intuitive Cue–Action Relationships}
Implicit control should be used when the relationship between sensed user behavior and robot action is predictable and easily understood by the user. In WAFFLE, participants felt most in control when the robot’s behavior aligned consistently with behaviors they already understood as meaningful, such as finishing chewing or pausing during conversation. Meanwhile, of the participants who preferred the adaptive strategy in the CRAFTT study, only some explicitly recognized the relationship between their reactions and the robot's behavior. Others described the interaction as intuitive without further reasoning, unable to clearly articulate why it felt that way. Notably, both systems did not rely on learned cues from the user; they used signals that were already embedded in how participants regulated eating and social interaction, in the case of WAFFLE, or how people naturally emote and express during interactions, for CRAFTT.

This alignment with natural user behavior supports predictability and intuitiveness in the interaction since users can form stable expectations, whether consciously or unconsciously, about how the robot will act. However, when sensing of these cues becomes misaligned, as we occasionally observed with missed chewing detection or unexpected pauses in WAFFLE, users could no longer accurately predict the robot's behavior, and their sense of control diminished. 
\textbf{Designers should selectively apply implicit control in situations where robot actions can be clearly and reliably linked to recognizable human behaviors.}

\subsection{How to Design for Effective Implicit Control}
\subsubsection{Reducing Workload}
Reducing cognitive and physical workload is an intrinsic advantage of automating a task, so this may appear to be something designers get ``for free'' when developing a system for implicit control. Our thematic analyses revealed that while this might be true on the whole, automating a cognitively or physically burdensome task can actually introduce new types of burden. When the robot's behavior is not aligned with the user's expectations, they may feel the need to be vigilant and more actively monitor the robot to ensure it doesn't do something unwanted or unsafe. Alternatively, if the robot is insufficiently responsive, users may implicitly compensate by adjusting their own actions, effectively taking on additional coordination effort. This suggests that \textbf{it is insufficient to simply eliminate direct control effort to reduce workload; users must find the system intuitive, otherwise they will feel they must compensate with increased attention to ensure the robot behaves appropriately.} 

\subsubsection{Preserving Sense of Control}
We found that, across WAFFLE and CRAFTT, users' sense of control over the robot was influenced by the responsiveness and intuitiveness, capacity to adapt to preferences, and contextual awareness of the system. Our analysis of CRAFTT specifically revealed that modulating the robot's behavior based on implicit cues from the user may have contributed to anthropomorphic impressions of the robot as ``understanding'' (CP1, CP4) and ultimately compliant to them, suggesting that developing features of the robot that dynamically adapt to the user is one avenue through which designers can allow users to feel in control. However, while adaptation can enhance perceived collaboration and intuitiveness, it may make the system feel less interpretable to some users. For example, some participants preferred the non-adaptive case over CRAFTT as they saw it as more consistent. We saw the same sentiment echoed by some participants in the WAFFLE study, with the main factor that negatively influenced the user's perception of their sense of control being unreliability of the system; if the robot misdetected a cue from the user and subsequently behaved incorrectly, this made users feel that the interaction was unnatural, that they could not accurately predict when and why the robot would act, and/or that the robot was not socially appropriate.

This finding suggests two things to designers. \textbf{First, implicit control systems should balance dynamic adaptation with predictability. Second, there is some minimum threshold of reliability that a system must reach for it to be perceived as sufficiently responsive and intuitive to enable users to feel in control, and therefore for the implicit paradigm interaction to be effective.} Notably, what exactly this minimum threshold is depends on both the target population and the needs of the individual user~\cite{bhattacharjee2020, wald2024mistakes}. For example, able-bodied users may be less tolerant of certain types of errors from a robotic system if they are physically capable of doing the task themselves, even if there may be an associated burden with doing so. On the other hand, the sense of control of users with impairments, who may need to rely on a human caregiver for the task in the absence of a robot, may still feel in control of the system even if there is some unreliability.

\section{Conclusion}
In our exploration of interaction-centric autonomous assistive robot development using WAFFLE and CRAFTT as design cases, we performed a thematic analysis of user feedback on interactions with these implicit control systems. Our analysis revealed that workload and sense of control were key factors that shaped user perception of these systems, with their sense of control during the interaction stemming from the intuitiveness and responsiveness, personalization, and contextual awareness of the robot. Notably, with WAFFLE, users reported a similar reduction of workload between WAFFLE and a fixed-interval feeding baseline, but ranked higher when it came to preserving users' sense of control. Similarly, CRAFTT was preferred over a non-adaptive baseline and also elicited responses associated with feeling in control. These responses highlight how the interaction-centric perspective of implicit control systems distinguishes them from those that are simply autonomous. From this, we find that there are two essential design conditions that an implicit control system must meet to be effective: 1) lowered user burden and 2) maintained or enhanced perceived control.

We propose a set of design guidelines for implicit control systems, first recognizing that selecting implicit control for a particular application is itself a design decision. Prior work has explored interaction paradigms such as active control, where users explicitly command robot actions, and shared control, where control is arbitrated between the user and the robot. We position implicit control, where natural cues from the user indirectly control robot behavior, as a complement to these. In this paradigm, implicit control is best suited for situations where other forms of control cause significant user burden, like in highly repetitive or socially sensitive scenarios, or where the user is vulnerable during the interaction. Further, the user's sense of control is mediated in part by their ability to accurately intuit how the robot will behave, so an effective implicit control system should rely on signals that are clearly interpretable. 
Our guidelines conclude by detailing key considerations for developing an implicit control system such that it is effective.

% \textcolor{red}{Importantly, both WAFFLE and CRAFTT illustrate that implicit control is not inherently tied to a specific task, but rather to a class of interaction characteristics. The principles we outline are most applicable in scenarios that are coordination-heavy, involve intimate interaction, and where explicit control would introduce significant cognitive or physical burden. In such contexts, both timing-based and execution-level forms of implicit control can be leveraged to regulate different aspects of interaction while preserving user comfort and sense of agency. These characteristics extend beyond assistive feeding to other assistive domains such as bathing, dressing, or rehabilitation, where interaction often involves both coordination of action timing and modulation of how actions are performed (e.g., force or speed). While the specific implementation may vary across tasks, the underlying design principles of reducing user burden while maintaining or enhancing perceived control remain consistent. This suggests that implicit control represents a task-agnostic design paradigm that can generalize across a range of assistive interactions.}

Based on our analysis, we contend that designing autonomous systems from an interaction-centric perspective, framed as implicit control, can meaningfully and positively impact users' experiences with these systems by preserving their sense of control. Importantly, we note that both WAFFLE and CRAFTT illustrate that implicit control is not inherently tied to a specific task, but rather to a class of interaction characteristics. While we examined this idea in an assistive context, our evaluations with both able-bodied individuals and those with impairments suggest that the core design conditions for effective implicit control ring true for both groups, although different populations, or even specific individuals, may personally weigh these conditions differently. We encourage designers to consider that, in practice, the same system may utilize implicit control at more than one level (e.g., for both bite timing and managing comfort in an integrated assistive feeding system) or may use it alongside active or shared control elements (e.g., an explicit safety override).
% (e.g., some users may value low workload over sense of control, others may have lower tolerance for unpredictable behavior from the robot), underscoring the need for customization and evaluation across measures with the target users. 
Overall, we recognize the under-utilization of interaction-centric paradigms in the design of autonomous systems and urge designers and researchers to consider these guidelines as they develop systems across contexts.

\section*{Acknowledgments}
We extend our gratitude to Pragathi Praveena for her guidance and feedback on scoping and narrative. We also thank our additional collaborators at both Carnegie Mellon and Cornell University for their contributions to the development of the original WAFFLE system that informed this work.

\bibliographystyle{IEEEtran}
\bibliography{bibliography}
\end{document}